\documentclass[11pt, a4paper, logo]{lumia}

\usepackage[numbers,sort&compress]{natbib}
\definecolor{infrecred}{HTML}{D2071E}
\definecolor{abstrabg}{HTML}{FCDDDC}
\colorlet{abstradark}{infrecred}
\definecolor{reportaccent}{HTML}{BB0819}
\colorlet{reporttoc}{reportaccent}
\definecolor{cayenne}{HTML}{941100}
\hypersetup{allcolors=cayenne}
\definecolor{darkblue}{rgb}{0, 0, 0.5}
\definecolor{lightblue}{HTML}{D0CFEA}

\usepackage{xspace}

\theoremstyle{plain}

\newtheorem*{proposition*}{Proposition}

\theoremstyle{definition}

\theoremstyle{definition}

\def\eqref#1{equation~\ref{#1}}

\usepackage{xurl}
\graphicspath{{figures/}}

\ifdefined\XeTeXversion\microtypesetup{tracking=false}\fi
\usepackage{colortbl}
\newcommand{\bsm}{\mathrm{BSM}}

\newif\ifincludeappendix
\includeappendixtrue

\usepackage[capitalize]{cleveref}
\crefname{section}{Sec.}{Secs.}
\Crefname{section}{Section}{Sections}
\crefname{table}{Tab.}{Tabs.}
\Crefname{table}{Table}{Tables}
\setheadertext{Representation Alignment Is Only as Good as the Teacher's Behavioral Structure}
\setheadertitle{Representation Alignment Is Only as Good as the Teacher's Behavioral Structure}
\title{Representation Alignment Is Only as Good as the Teacher's Behavioral Structure}
\author{\parbox{\textwidth}{\centering
  {\Authfont Zhengfeiyang~Zhang$^{1}$ \quad Haoyuan~Chen$^{2}$ \quad
  Hexin~Wang$^{3}$ \quad Hanyu~Zhou$^{4}$\par
  Zhenglin~Wan$^{4}$ \quad Wangbo~Zhao$^{5}$ \quad Pengfei~Zhou$^{4,\ddagger}$\par}
  \vspace{0.35em}
  {\Affilfont
   $^{1}$The University of Tokyo \quad $^{2}$University of the Chinese Academy of Sciences \quad $^{3}$InfRec\par
   $^{4}$National University of Singapore \quad $^{5}$The Hong Kong University of Science and Technology\par}}}
\correspondingemail{$^{\ddagger}$Corresponding author: \href{mailto:zpf4wp@outlook.com}{zpf4wp@outlook.com}}

\begin{document}
\begin{abstract}
Recent works align the intermediate features of vision-language-action (VLA) models with frozen vision encoders during fine-tuning, hoping to inject additional priors such as 3D awareness or affordances. Unlike in image generation, where teachers for representation alignment (REPA) are assessed with encoder-intrinsic metrics, teachers for VLAs are still chosen largely by heuristics. We show that this practice is unreliable: across 14 teachers and 16 manipulation tasks, under matched training budgets, no off-the-shelf vision encoder improves the clean success rate of $\pi_{0.5}$ by more than 1 point, and all of them reduce its success rate under perturbations by 10 to 39 points. Neither global semantics nor spatial structure, the criteria used in image generation, predicts which teachers help. We therefore propose the behavioral structure metric (BSM), which measures whether states that are close in a teacher's feature space are followed by similar expert actions within the same task. BSM correlates strongly with success rate under perturbations ($r=0.87$), and every off-the-shelf vision encoder scores below $\pi_{0.5}$ itself. Based on it, we introduce Behavior-Aware VLA Alignment (BAVA), a general VLA alignment protocol that favors vision encoders whose BSM exceeds the student's and stops alignment early according to the student's BSM during training. On SmolVLA, BAVA matches continuous alignment with one third of the alignment budget, and improves the average success rate by 4.6 points over unaligned fine-tuning. Our results suggest that representation alignment benefits VLAs only when the teacher encodes rich behaviorally relevant structure.
\end{abstract}

\renewcommand{\absfont}{\linespread{1.05}\fontsize{10}{12}\selectfont}
\teaserfigure{%
\centering
\includegraphics[width=\linewidth]{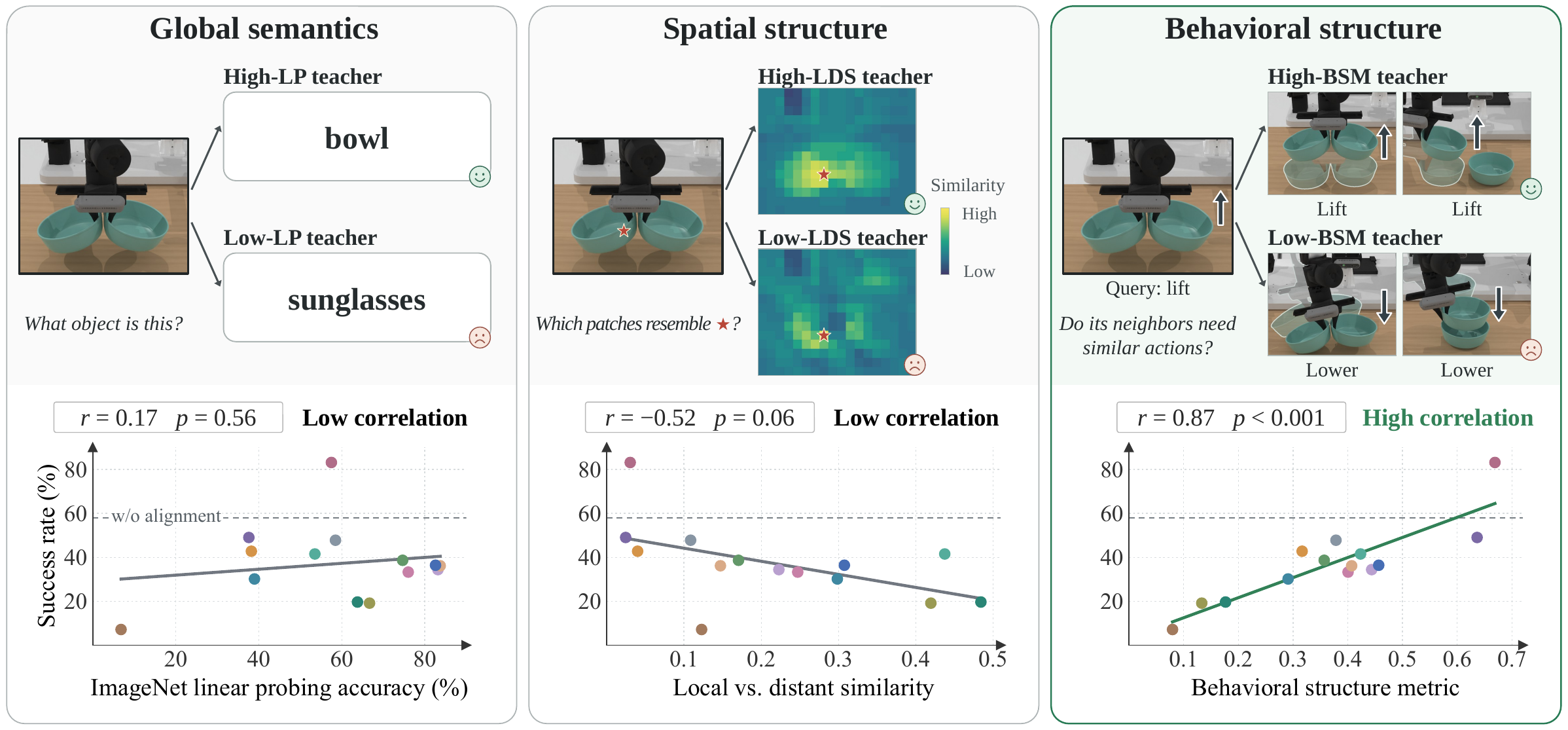}\par
{\captionsetup{font=footnotesize,skip=4pt}
  \captionof{figure}{\textbf{What makes a good teacher for a VLA?}
    \textit{Top:} Three candidate metrics, each illustrated on the same observation with a high-scoring and a low-scoring teacher.
    Linear probing (LP) asks whether a linear classifier on the teacher's features recognizes the object; local vs.\ distant similarity (LDS) asks whether patches near the starred patch are more similar to it than distant ones; our behavioral structure metric (BSM) asks whether the observation's neighbors in the teacher's feature space are followed by similar expert actions.
    \textit{Bottom:} Success rate under perturbations for a $\pi_{0.5}$ student aligned with 14 teachers.
    Neither LP nor LDS predicts whether alignment helps, whereas BSM does.}
  \label{fig:teaser}
}
}
\maketitle
\clearpage
\begingroup
\hypersetup{linkcolor=reporttoc,linktoc=section}
\setlength{\parskip}{0pt}
\setcounter{tocdepth}{3}
\tableofcontents
\endgroup
\clearpage
\section{Introduction}
\label{sec:introduction}
Vision-language-action (VLA) models~\cite{kim2024openvla,black2025pi0,pi2025pi05} inherit visual representations from vision-language pretraining and are then fine-tuned through behavior cloning. A growing body of work adds representation alignment during fine-tuning: intermediate visual representations from one or more layers of a VLA (the student) are aligned with features from an external frozen vision encoder (the teacher) during training, and the teacher is removed at inference time. Some methods aim to inject spatial, geometric, or affordance priors that typical VLAs may lack~\cite{li2026spatialforcing,zhu2026rigid,sun2026rocket,kong2026affordvla}; others seek to preserve the generalization acquired during pretraining by constraining changes to pretrained representations~\cite{kachaev2026blindvla,huang2025maps}. The idea comes from diffusion models, where representation alignment (REPA) showed that aligning a denoising transformer to a pretrained encoder speeds up training and improves generation quality~\cite{yu2025repa}.

In diffusion models, which teacher to align to is a central question, and it is answered with metrics computed on the teacher alone, before any training. Early work suggested that teachers with stronger global semantics, measured by linear probing accuracy, are better targets~\cite{yu2025repa}; later work evaluated the spatial structure among patch tokens and found it to be a stronger predictor of generation quality~\cite{singh2026irepa}. For VLAs, however, teachers are still largely selected because their features appear to encode useful priors for manipulation, such as depth or 3D geometry. Whether these properties make a teacher useful for action learning is typically assessed only after alignment~\cite{li2026spatialforcing}, through success rates on nearly saturated benchmarks~\cite{liu2023libero}.

We argue that this practice obscures what alignment actually contributes. VLA fine-tuning differs from image generation in two key respects. First, REPA supervises a diffusion transformer trained from scratch, providing learned visual structure that is absent at random initialization. A VLA, however, already has pretrained visual representations, raising the question of whether an external vision encoder offers stronger representations for action learning. Second, a policy needs more than geometry or semantics in the general sense: it must separate states that call for different actions even when they look alike, and group states that call for the same action even when they look different. The usefulness of off-the-shelf vision encoders as alignment targets therefore needs to be reassessed against the requirements of action learning.

To investigate this question, we conduct a controlled study of 14 teachers. Eleven are off-the-shelf vision encoders spanning image self-supervision, image--text alignment, segmentation and object understanding, 3D geometry, multi-teacher distillation, video self-supervision, and embodied visual control. The other three provide reference points: a randomly initialized vision transformer (ViT)~\cite{dosovitskiy2020vit} without learned visual priors, a $\pi_0$~\cite{black2025pi0} policy fine-tuned on the same tasks, and a $\pi_{0.5}$~\cite{pi2025pi05} policy trained on expert demonstrations from perturbed environments, which serves as an oracle because the student never sees such data. Beyond teacher selection, the study examines four ways to impose alignment: single-layer patch-to-patch alignment, attention-map alignment, multi-layer feature alignment, and per-camera global feature alignment. Evaluation covers 16 representative RoboTwin~\cite{chen2025robotwin2} tasks spanning precision contact, pick-and-place, bimanual, and multi-stage and articulated manipulation, in a Clean setting that matches the training environments and a Randomized setting that adds distractors and randomizes textures and lighting. Together, these experiments lead to \emph{three findings and a recipe}:

\noindent\textbf{Small gains, substantial harm.} With $\pi_{0.5}$ as the student, no off-the-shelf vision encoder improves the final success rate in the Clean setting by more than 1 point, while all of them lower it in the Randomized setting, by 10 to 39 points. A close inspection of the rollouts points to reduced manipulation precision as the main source of failure.

\noindent\textbf{Teacher-selection criteria from image generation do not transfer to VLAs.} We examine the two criteria studied for REPA: global semantics, measured by ImageNet~\cite{deng2009imagenet} linear probing (LP) accuracy, and spatial structure, measured by local vs.\ distant similarity (LDS). Across the 14 teachers, both metrics show only a weak linear correlation with the success rate of the aligned policy.

\noindent\textbf{The teacher's behavioral structure matters.} We propose the behavioral structure metric (BSM), which measures whether states that are close in a teacher's feature space are followed by similar expert actions (\cref{fig:teaser}). BSM shows a strong linear correlation with the success rate of the aligned policy ($r=0.87$). Every off-the-shelf vision encoder has a lower BSM than $\pi_{0.5}$ itself, so alignment pulls the student toward weaker behavioral structure than its own. Conversely, the oracle, whose BSM exceeds the student's, is the only teacher that improves $\pi_{0.5}$, and DINOv2~\cite{oquab2024dinov2} harms $\pi_{0.5}$ but improves OpenVLA-OFT~\cite{kim2024openvla,kim2025oft}, whose BSM is lower than DINOv2's.

\noindent\textbf{From diagnosis to recipe.} These findings lead to Behavior-Aware VLA Alignment (BAVA): favor a teacher whose BSM is higher than the student's, and stop alignment early according to the student's BSM during training. On SmolVLA~\cite{shukor2025smolvla}, BAVA matches continuous alignment with one third of the alignment budget, and improves the average success rate by 4.6 points over unaligned fine-tuning.

\section{Setup}
\label{sec:setup}
\subsection{Representation Alignment in VLAs}
Let $H^\ell$ denote the student visual tokens at layer $\ell$, and $T^m$ the teacher features at layer $m$ for the same observation. Representation alignment adds an alignment term to the behavior cloning loss. For a selected set of layer pairs $\mathcal{P}=\{(\ell,m)\}$, readout functions $\phi_S$ and $\phi_T$ extract the objects to be aligned from the corresponding student and teacher layers. An alignment operator $\mathcal{A}$ measures their discrepancy. The terms are summed with weights $w_{\ell m}$ and scaled by a coefficient $\lambda(t)$ that varies with training step $t$:
\begin{equation}
\mathcal{L}=\mathcal{L}_{\text{behavior cloning}}+\lambda(t)
\sum_{(\ell,m)\in\mathcal{P}}w_{\ell m}\,
\mathcal{A}\bigl(\phi_S(H^\ell),\phi_T(T^m)\bigr).
\label{eq:alignment}
\end{equation}
Existing methods are special cases of this formulation. They differ in the teacher, layer pairs $\mathcal{P}$ and weights $w$, readouts $\phi$, alignment operator $\mathcal{A}$, and schedule $\lambda(t)$. For example, Spatial Forcing~\cite{li2026spatialforcing} aligns a single layer pair ($|\mathcal{P}|=1$) patch by patch; $\phi_S$ is a learnable projection head, and $\mathcal{A}$ averages $1-\cos$ over tokens. The AffordVLA~\cite{kong2026affordvla} teacher takes both images and instructions as input, with $\phi_T$ producing task-relevant affordance features. RIGID~\cite{zhu2026rigid} aligns attention maps, while ROCKET~\cite{sun2026rocket} aligns multiple layer pairs simultaneously ($|\mathcal{P}|>1$). When comparing teachers, we fix the student and alignment configuration and vary only the teacher. When reproducing other methods, we use the configurations specified in their respective papers.

\subsection{Teachers and Alignment Methods}
\noindent\textbf{Teachers.} We evaluate 14 teachers. Eleven are off-the-shelf vision encoders from seven families: image self-supervision (DINOv2~\cite{oquab2024dinov2} and MAE~\cite{he2022mae}), image--text alignment (SigLIP~\cite{zhai2023siglip}), segmentation and object understanding (SAM3~\cite{carion2026sam3}), 3D geometry (VGGT~\cite{wang2025vggt} and DA3~\cite{lin2026da3}), multi-teacher distillation (RADIO~\cite{ranzinger2024radio}), video self-supervision (V-JEPA 2.1~\cite{murlabadia2026vjepa21} and DINO-Video~\cite{wu2026lingbotvla2}), and embodied visual control (CAIP~\cite{sharma2026caip} and VC-1~\cite{majumdar2023vc1}). The remaining three are reference teachers: Random ViT, a randomly initialized vision transformer; a $\pi_0$ policy fine-tuned on the same 16 tasks; and the oracle, a $\pi_{0.5}$ policy trained on expert demonstrations from the Randomized setting, which the student never sees.

\noindent\textbf{Alignment methods.} We compare four alignment methods that differ in the representations they constrain. Patch-to-patch alignment~\cite{li2026spatialforcing} uses a projection head and cosine loss to match student features from one intermediate layer to teacher features at corresponding spatial positions. Attention-map alignment~\cite{zhu2026rigid} matches the student's learned query--key relations to the teacher's native attention through a spatial Sinkhorn optimal transport objective. Multi-layer alignment~\cite{sun2026rocket} averages patch-wise cosine losses over 10 teacher--student layer pairs using a shared Matryoshka projection head. Global alignment independently mean-pools valid patches from the head, left-wrist, and right-wrist cameras, then applies projection and cosine alignment to the pooled features, yielding one global constraint per view. Unless otherwise specified, we use patch-to-patch alignment by default, as it is widely adopted in both image generation and VLAs.

\subsection{Tasks, Evaluation, and Training}
\noindent\textbf{Students and data.} Our primary student is $\pi_{0.5}$~\cite{pi2025pi05}, fine-tuned on 16 RoboTwin tasks using 50 Clean demonstrations per task, with alignment at the 12th block of the vision-language model (VLM) by default. Teachers receive unaugmented $224\times224$ inputs. The student uses only color augmentation, preserving exact token correspondence. We also use $\pi_0$~\cite{black2025pi0}, OpenVLA-OFT~\cite{kim2024openvla,kim2025oft}, and SmolVLA~\cite{shukor2025smolvla} as students.

\noindent\textbf{Evaluation.} We evaluate under two settings, \emph{Clean} and \emph{Randomized}, with 800 episodes each. Clean uses unperturbed environments, whereas Randomized combines five distractors per episode with randomized textures and lighting, following the domain randomization protocol of RoboTwin~2.0~\cite{chen2025robotwin2}.

\noindent\textbf{Training.} We observe that training of $\pi_{0.5}$ converges after approximately 20k iterations. To ensure a fair comparison under a consistent training budget, we train all runs for 30k iterations unless otherwise specified.

\section{Motivation: Alignment Falls Short of Expectations}
\label{sec:motivation}
\begin{figure}[htbp]
  \centering
  \includegraphics[width=.79\linewidth]{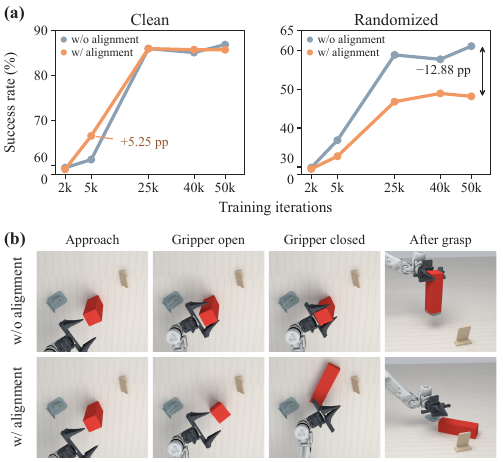}
  \caption{\textbf{Early gains do not translate into robust manipulation.}
    (a) Success rates of $\pi_{0.5}$ with and without patch-to-patch alignment to VGGT, as in Spatial Forcing~\cite{li2026spatialforcing}. Each point averages 800 episodes over 16 tasks. In the Clean setting, alignment helps early (+5.25~pp at 5k iterations), but the gain vanishes by 25k. In the Randomized setting, it lowers the final success rate by 12.88~pp.
    (b) Both policies on the same Randomized block-handover episode, with columns matched by grasp phase. Without alignment, the policy centers its open gripper around the block and lifts it. With alignment, the open gripper tips the block over and then closes beside it, and the grasp fails. For visibility, the recorded trajectories are re-rendered with a muted table texture, from an overhead view (first three columns) and a side view (last column).}
  \label{fig:overview}
\end{figure}

We begin with a concrete example: our reproduction of Spatial Forcing on $\pi_{0.5}$ using VGGT as the teacher. To examine whether early gains persist with sufficient training, we train both the aligned policy and the unaligned baseline for 50k iterations. As shown in \cref{fig:overview}, alignment improves success rate in the Clean setting by 5.25 percentage points (pp) at 5k, but this advantage vanishes as training continues. In the Randomized setting, alignment instead reduces success rate by 12.88~pp at 50k. A close inspection of execution videos shows that the vast majority of failures involve positioning errors during grasping. Ironically, the 3D teacher was introduced to improve spatial understanding and enable more precise manipulation, yet the aligned policy becomes less precise and less reliable under perturbations.

This counterintuitive observation prompts a closer examination of teacher selection. A teacher may encode useful geometric information, yet aligning to its features need not improve the policy. An essential question then arises: which properties of a teacher predict whether alignment helps a VLA?

\begin{figure}[tbp]
  \centering
  \includegraphics[width=.74\linewidth]{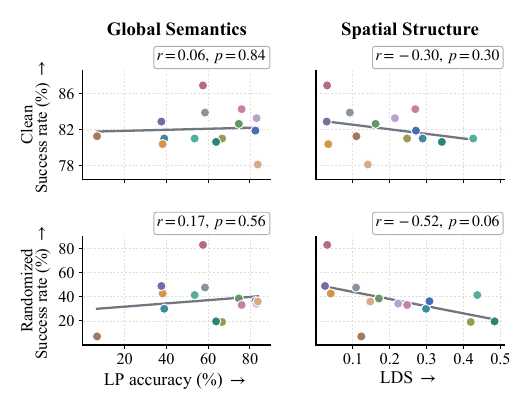}
  \caption{\textbf{LP and LDS do not predict teacher quality.} Each point is one teacher, plotted by its LP accuracy or LDS against the success rate of the $\pi_{0.5}$ student aligned to it (same setup as \cref{fig:teaser}; colors denote teachers). Gray lines show least-squares fits across the 14 teachers; boxes report Pearson's $r$ and its two-sided $p$-value. Neither metric correlates significantly with success rate in the Clean setting (top) or the Randomized setting (bottom).}
  \label{fig:lp_lds_outcomes}
\end{figure}

We first examine two teacher-selection criteria studied in representation alignment for diffusion transformers: ImageNet LP accuracy and LDS~\cite{yu2025repa,singh2026irepa}. Across the 14 teachers, we ask whether higher scores predict a higher success rate for $\pi_{0.5}$ after 30k iterations of alignment (\cref{fig:lp_lds_outcomes}).

\textbf{Stronger global semantics do not guarantee a better teacher.} We train a linear classifier on mean-pooled patch features from each frozen teacher and measure top-1 accuracy on ImageNet-1K~\cite{deng2009imagenet}. LP accuracy shows no statistically significant correlation with policy success in either the Clean ($r=0.06$) or Randomized ($r=0.17$) setting. For example, DA3 and CAIP differ by less than 1~pp in probe accuracy, yet the policies aligned to them differ by 12.6~pp in Randomized success rate (42.75\% versus 30.13\%).

\textbf{Stronger spatial structure does not guarantee a better teacher.} We compute LDS by subtracting the mean cosine similarity of distant patch pairs from that of nearby pairs, using a Manhattan-distance threshold of $H/2$ on an $H\times H$ patch grid. Within each setting, we use the same observations for all teachers and average across the three camera views. The fitted trend is even negative in both the Clean ($r=-0.30$) and Randomized ($r=-0.52$) settings, though neither correlation is statistically significant. For example, SAM3 has a much higher Randomized LDS than VGGT (0.44 versus 0.11), yet aligning to SAM3 yields a lower success rate (41.50\% versus 47.75\%).

\textbf{Toward a criterion tailored to VLA policies.} These results raise a further question: what makes a teacher's representation beneficial for action prediction, and how can we assess that property before alignment? Following the goal of diagnosing teacher suitability pursued by iREPA for diffusion transformers~\cite{singh2026irepa}, we turn to the behavioral structure of representations and ask whether they organize observations in a way that is consistent with expert behavior.

\section{Behavioral Structure Matters}
\label{sec:behavioral_structure}
\subsection{Hypothesis: Policies Need Behavioral Neighborhoods}
Consider the grasp in \cref{fig:overview}. Changing the table texture can leave the required grasp unchanged, while shifting the block a few centimeters can require the gripper to approach from a different position. A teacher can recognize the block and preserve local patch structure in every image, yet still miss the offset between the block and the gripper.

\begin{figure}[t]
  \centering
  \includegraphics[width=\linewidth]{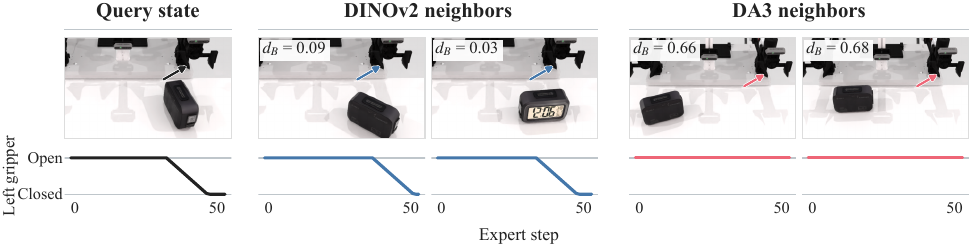}
  \caption{\textbf{DINOv2 retrieves states that require the same action; DA3 does not.} For a query state from the Clean RoboTwin alarm-clock task, we show its two nearest neighbors in each teacher's feature space, all at the retrieval step. Arrows mark the left gripper, and traces plot its opening over the next 50 expert steps. The gripper closes in the query and in both DINOv2 neighbors, but stays open in both DA3 neighbors. $d_B$ is the distance between the neighbor's and the query's two-arm action sequences (\cref{sec:bsm}, step~\ref{step:behavioral_distance}); lower means more similar.}
  \label{fig:bsm_retrieval}
\end{figure}

Prior work on visual control suggests judging state similarity by what the agent can do. One approach groups states when taking the same action leads to similar rewards and subsequent states~\cite{zhang2021learning}. Another brings states closer when an optimal policy behaves similarly now and in the steps that follow~\cite{agarwal2021pse}. For a VLA trained through behavior cloning, demonstrations give us a direct way to examine this idea: when a frozen teacher places two observations from the same task close together, does the expert take similar actions over the next few steps?

We therefore hypothesize that \textbf{a teacher is a more effective alignment target when nearby states in its feature space have similar expert action sequences.} We test whether, for a fixed student and alignment configuration, teachers with higher BSM scores yield better downstream performance, and whether BSM has an approximately linear relationship with the change in success rate after alignment. We further examine whether this predictive relationship holds across alignment methods and students.

\subsection{Behavioral Structure Metric}
\label{sec:bsm}
\label{sec:bsm_prediction}
BSM measures whether states that are close in the teacher's feature space also have similar expert actions. Intuitively, if we use teacher features to retrieve the demonstration states most similar to the current state, the expert actions in those states should agree with the current state's action. Greater agreement means that the teacher features better reflect differences in actions. \Cref{fig:teaser} illustrates this intuition, and \cref{fig:bsm_retrieval} shows the actions that follow a query and its retrieved neighbors in an expert demonstration dataset.

Let $\mathcal{T}$ be the set of tasks and $\mathcal{S}_\tau$ the states from task $\tau$. The computation has three steps:
\begin{enumerate}
  \item \textbf{Retrieve feature neighbors.} For each state $i\in\mathcal{S}_\tau$, we use pooled teacher features to retrieve its $k$ nearest neighbors among states $j\in\mathcal{S}_\tau$ with $j\ne i$, denoted by $\mathcal{N}_k^\tau(i)$.
  \item\label{step:behavioral_distance} \textbf{Measure behavioral distance.} For each neighbor $j$, we compute its behavioral distance from $i$, $d_B(i,j)$, using the next $h$ steps of expert actions after each state. These sequences comprise joint positions relative to the current state and gripper openings.
  \item \textbf{Normalize against random neighbors.} We normalize the mean behavioral distance to retrieved neighbors by the expected distance to states $j$ sampled uniformly from $\mathcal{S}_\tau$ with $j\ne i$. We average the normalized distances over states within each task and subtract from one, then average the resulting scores equally across tasks:
\end{enumerate}
\begin{equation*}
\bsm=\frac{1}{|\mathcal{T}|}\sum_{\tau\in\mathcal{T}}\left[
1-\frac{1}{|\mathcal{S}_\tau|}\sum_{i\in\mathcal{S}_\tau}
\frac{\frac{1}{k}\sum_{j\in\mathcal{N}_k^\tau(i)}d_B(i,j)}
{\mathbb{E}_{j\in\mathcal{S}_\tau,\,j\ne i}\,d_B(i,j)}
\right].
\end{equation*}

\begin{table}[t]
  \centering
  \caption{\textbf{Correlation between teacher metrics and aligned-student success rate.} We report Pearson's $r$ and two-sided $p$-values across 14 teachers, using the success rate of the aligned $\pi_{0.5}$ student at 30k iterations. For \emph{Average}, each teacher's metric score and success rate are first averaged over the Clean and Randomized settings and then correlated across teachers. Only BSM correlates significantly with success rate, in the Randomized and Average columns.}
  \label{tab:metric_correlations}
  \small
  \setlength{\tabcolsep}{2pt}
  \begin{tabular*}{.76\linewidth}{@{}l@{\extracolsep{\fill}}rr@{\hspace{16pt}}rr@{\hspace{16pt}}rr@{}}
    \toprule
    & \multicolumn{2}{c}{Clean} & \multicolumn{2}{c}{Randomized} & \multicolumn{2}{c}{Average} \\
    \cmidrule(lr){2-3}\cmidrule(lr){4-5}\cmidrule(l){6-7}
    Metric & $r$ & $p$ & $r$ & $p$ & $r$ & $p$ \\
    \midrule
    LP  & $0.06$  & $0.837$ & $0.17$  & $0.562$  & $0.16$  & $0.575$ \\
    LDS & $-0.30$ & $0.301$ & $-0.52$ & $0.056$  & $-0.49$ & $0.073$ \\
    BSM & $0.49$  & $0.073$ & {\boldmath$0.87$} & {\boldmath$\vcenter{\hbox{\scriptsize\boldmath$<$}}\,0.001$} & {\boldmath$0.80$} & {\boldmath$\vcenter{\hbox{\scriptsize\boldmath$<$}}\,0.001$} \\
    \bottomrule
  \end{tabular*}
\end{table}

We use $k=10$ neighbors and an action horizon of $h=50$ steps, and compute BSM on 1,600 states per setting (100 per task). BSM is upper-bounded by 1. Values closer to 1 indicate that feature neighbors have more similar expert actions, while values near 0 indicate action similarity close to that of random states from the same task. Negative values occur when retrieved neighbors have larger normalized behavioral distances than random states, on average.

\textbf{BSM predicts alignment outcomes.} Among the three metrics, only BSM tracks how well the aligned student performs (\cref{tab:metric_correlations}). In the Randomized setting, a teacher's BSM correlates strongly with the student's success rate ($r=0.87$), whereas neither LP accuracy nor LDS shows a significant correlation across the same teachers (\cref{fig:lp_lds_outcomes}). The BSM correlation remains significant when each teacher's score and the corresponding student's success rate are averaged across both settings. In the Clean setting, the correlation is weaker and does not reach significance, though it is still the strongest of the three. Clean success rates span only 78--87\% across teachers, compared with 7--83\% under randomization, leaving less variation for any metric to explain. These results suggest that a teacher's value for VLA alignment depends on how well its features group states requiring similar actions. Neither global semantics nor spatial structure captures this relationship.

All evaluated off-the-shelf vision encoders have lower BSM than the pretrained $\pi_{0.5}$ (0.591 in the Clean setting and 0.497 in the Randomized setting), and none improves its final success rate by more than 1 point. Together, these observations suggest that a teacher's BSM should be judged relative to the student's existing behavioral structure.

We next examine whether VLAs with lower BSM can benefit from alignment to teachers with higher BSM than their own.

\subsection{Generalization across Students and Alignment Methods}
\label{sec:consistency}
\noindent\textbf{Different students.}
To test whether our findings extend beyond $\pi_{0.5}$, we repeat the comparison with two other students, aligning each patch-to-patch (\cref{tab:oft_alignment}). For OpenVLA-OFT, both DINOv2 and DA3 have a higher BSM than the pretrained student, and both improve over unaligned fine-tuning. DINOv2, whose BSM exceeds the student's by 0.195, raises the average success rate by 8.25~pp, whereas DA3, with a lead of only 0.026, raises it by 2.19~pp. For $\pi_0$, neither DINOv2 nor VGGT has a higher BSM than the pretrained student, and both lower the average success rate, by 4.69 and 4.88~pp.

\begin{table}[t]
  \centering
  \caption{\textbf{Alignment outcomes on OpenVLA-OFT and $\pi_0$.}
    BSM is averaged over the Clean and Randomized settings; the first row
    of each block reports the BSM of the pretrained student and the success rates of
    unaligned fine-tuning, and $\Delta$ denotes the difference from that row.
    Both teachers with a higher BSM than OpenVLA-OFT improve its success rate, and both teachers with a lower BSM than $\pi_0$ reduce it.}
  \label{tab:oft_alignment}
  \small
  \setlength{\tabcolsep}{2pt}
  \begin{tabular*}{.86\linewidth}{@{\extracolsep{\fill}}lrrrr@{}}
    \toprule
    & & \multicolumn{3}{c}{Success rate (\%)} \\
    \cmidrule(l){3-5}
    Alignment target & BSM ($\Delta$) & Clean & Randomized & Average ($\Delta$) \\
    \midrule
    \multicolumn{5}{@{}>{\columncolor{gray!15}[0pt][0pt]}l@{}}{\emph{OpenVLA-OFT}} \\
    w/o alignment & 0.314 & 33.00 & 2.00 & 17.50 \\
    DA3 & 0.340 (+0.026) & 36.63 & 2.75 & 19.69 (+2.19) \\
    DINOv2 & 0.509 (+0.195) & 43.75 & 7.75 & 25.75 (+8.25) \\
    \midrule
    \multicolumn{5}{@{}>{\columncolor{gray!15}[0pt][0pt]}l@{}}{\emph{$\pi_0$}} \\
    w/o alignment & 0.522 & 72.50 & 20.63 & 46.56 \\
    DINOv2 & 0.509 ($-$0.013) & 71.50 & 12.25 & 41.88 ($-$4.69) \\
    VGGT & 0.414 ($-$0.108) & 69.25 & 14.13 & 41.69 ($-$4.88) \\
    \bottomrule
  \end{tabular*}
\end{table}

\noindent\textbf{Different alignment methods.}
With VGGT as the teacher and $\pi_{0.5}$ as the student, we compare four alignment methods: single-layer patch-to-patch alignment following Spatial Forcing, attention-map alignment following RIGID, multi-layer alignment following ROCKET, and global alignment of per-view pooled features (\cref{tab:methods}). None of the four improves over unaligned fine-tuning in either setting. The methods perform similarly in the Clean setting, and all of them lower the Randomized success rate: by 0.88~pp with global alignment and by 10.38 to 17.50~pp with the other three. This is consistent with VGGT's BSM being lower than the student's: changing how alignment is imposed does not make such a teacher useful. This ordering is intuitive: global alignment imposes the weakest constraint on the student, so it does the least harm. By the same reasoning, it should yield a smaller gain when the teacher is beneficial, which we confirm in \cref{tab:bava_ablation}.

\begin{table}[t]
  \centering
  \caption{\textbf{All four alignment methods lower the success rate of $\pi_{0.5}$ aligned to VGGT.}
    Success rates (\%) with four alignment methods.
    The methods perform similarly in the Clean setting; under randomization, global alignment causes the smallest drop.}
  \label{tab:methods}
  \small
  \begin{tabular*}{.70\linewidth}{@{\extracolsep{\fill}}lrrr@{}}
    \toprule
    Alignment method & Clean & Randomized & Average \\
    \midrule
    w/o alignment & 84.00 & 58.13 & 71.06 \\
    \midrule
    Patch-to-patch & 83.88 & 47.75 & 65.81 \\
    Attention map & 81.38 & 46.88 & 64.13 \\
    Multi-layer & 82.75 & 40.63 & 61.69 \\
    Global & 83.25 & 57.25 & 70.25 \\
    \bottomrule
  \end{tabular*}
\end{table}

\section{BAVA: Behavior-Aware VLA Alignment}
\label{sec:selection}
The findings in \cref{sec:behavioral_structure} suggest that effective alignment should draw on a teacher's richer behavioral structure to support the student's action learning. We translate this insight into Behavior-Aware VLA Alignment (BAVA), a simple recipe that determines which teacher to use and when to stop alignment.

\begin{table*}[t]
  \centering
  \caption{\textbf{Policy success rates across 16 RoboTwin tasks.} Each cell gives the mean success rate (\%) across the two settings, followed by (Clean / Randomized). The final column averages over all 16 tasks. $^\dagger$~Trained separately on each task; all other policies are trained jointly on the 16 tasks. Spatial Forcing is our reproduction on SmolVLA, following the original paper.}
  \label{tab:bava_simulation}
  \small
  \setlength{\tabcolsep}{2.5pt}
  \newcommand{\srpair}[3]{\begin{tabular}[t]{@{}c@{}}#1\\[-2pt]{\footnotesize(#2/#3)}\end{tabular}}
  \begin{tabular*}{\textwidth}{@{\extracolsep{\fill}}lccccc@{}}
    \toprule
    Method & Precision contact & Pick-and-place & Bimanual & Multi-stage \& articulated & Avg. \\
    \midrule
    DP$^\dagger$~\cite{chi2023diffusionpolicy} & \srpair{32.38}{63.50}{1.25} & \srpair{18.13}{36.25}{0.00} & \srpair{15.75}{31.50}{0.00} & \srpair{14.75}{29.50}{0.00} & \srpair{20.25}{40.19}{0.31} \\[2pt]
    ACT$^\dagger$~\cite{zhao2023act} & \srpair{34.50}{60.75}{8.25} & \srpair{25.50}{50.50}{0.50} & \srpair{30.75}{61.50}{0.00} & \srpair{18.38}{36.75}{0.00} & \srpair{27.28}{52.38}{2.19} \\[2pt]
    StarVLA~\cite{starvla2026} & \srpair{37.88}{73.50}{2.25} & \srpair{29.63}{57.50}{1.75} & \srpair{27.63}{50.00}{5.25} & \srpair{17.75}{34.25}{1.25} & \srpair{28.22}{53.81}{2.63} \\[2pt]
    SmolVLA~\cite{shukor2025smolvla} & \srpair{33.25}{52.00}{14.50} & \srpair{41.00}{73.00}{9.00} & \srpair{22.50}{40.00}{5.00} & \srpair{29.50}{52.50}{6.50} & \srpair{31.56}{54.38}{8.75} \\[2pt]
    Spatial Forcing~\cite{li2026spatialforcing} & \srpair{33.25}{47.00}{19.50} & \srpair{43.25}{74.00}{12.50} & \srpair{26.00}{49.00}{3.00} & \srpair{29.50}{52.50}{6.50} & \srpair{33.00}{55.63}{10.38} \\[2pt]
    \midrule
    \textbf{BAVA (ours)} & \srpair{37.50}{55.00}{20.00} & \srpair{48.75}{78.00}{19.50} & \srpair{27.25}{49.50}{5.00} & \srpair{31.25}{54.00}{8.50} & \srpair{\textbf{36.19}}{59.13}{13.25} \\
    \bottomrule
  \end{tabular*}
\end{table*}

\textbf{Favor teachers with substantially higher BSM than the student.} Before training, we evaluate candidate teachers and the VLA student on the same expert demonstrations using the BSM protocol in \cref{sec:bsm}, and favor teachers with a clear BSM advantage over the student. This criterion follows from \cref{sec:behavioral_structure}: $\pi_{0.5}$ and $\pi_0$ already have a higher BSM than every off-the-shelf vision encoder we evaluate, and alignment improves neither of them, whereas OpenVLA-OFT, whose BSM is lower, benefits from alignment. In this section, we apply BAVA to SmolVLA, another student with a low BSM, to further validate this recipe.

\textbf{Stop alignment once the student's BSM plateaus above the teacher's.} During fine-tuning, the student's BSM rises quickly, surpasses the frozen teacher's, and then levels off. Beyond this point, alignment no longer raises the student's BSM, and the teacher has no richer behavioral structure to offer. We therefore evaluate the student's BSM on a fixed set of held-out demonstrations every 5k iterations and turn off the alignment loss once it stops rising above the teacher's level. For SmolVLA, the student's BSM rises from 0.46 to 0.56 during the first 5k iterations, above the teacher's 0.51, and stays at 0.56 at 10k, so we stop alignment at 10k. Action learning continues after alignment stops. In image generation, stopping alignment at the right stage has likewise been shown to accelerate training and improve generation quality~\cite{wang2025haste}.

\subsection{Simulation Experiments}
\label{sec:bava_simulation}
\noindent\textbf{Experimental setting.} We train SmolVLA jointly on 16 RoboTwin tasks using 50 Clean demonstrations per task, fine-tuning its VLM together with the action expert. BAVA uses DINOv2 as the teacher, whose BSM (0.509) exceeds that of the pretrained SmolVLA (0.459) by 0.050, aligns patch to patch at the 11th VLM layer with $\lambda=0.1$, and turns off alignment at 10k iterations. We compare it with unaligned fine-tuning, with Spatial Forcing, which aligns to VGGT throughout training, and with two ablations that align to DINOv2 throughout training, either patch to patch or globally. All runs use the same training data, initialization, and 30k iterations. We evaluate each policy over 50 episodes per task in each of the Clean and Randomized settings (1,600 episodes per policy).

\noindent\textbf{Evaluation results.} BAVA raises the average success rate of SmolVLA from 31.56\% to 36.19\%, improving both the Clean (+4.75~pp) and Randomized (+4.50~pp) settings (\cref{tab:bava_simulation}). Spatial Forcing improves the average by 1.44~pp. \Cref{tab:bava_ablation} isolates the two choices in BAVA. With the same teacher, global alignment gains 1.31~pp and patch-to-patch alignment 4.31~pp. BAVA matches continuous patch-to-patch alignment while aligning for only the first 10k of 30k iterations.

\begin{table}[t]
  \centering
  \caption{\textbf{Ablation of BAVA on SmolVLA.} All aligned rows use DINOv2 as the teacher. Success rates (\%) are averaged over 800 episodes per setting. Avg. is the mean of Clean and Randomized success rates. The final row is BAVA. Bold marks the best result in each column.}
  \label{tab:bava_ablation}
  \small
  \setlength{\tabcolsep}{2pt}
  \begin{tabular*}{.78\linewidth}{@{\extracolsep{\fill}}lcccc@{}}
    \toprule
    \multicolumn{2}{c}{Configuration} & \multicolumn{3}{c}{Success rate (\%)} \\
    \cmidrule(r){1-2}\cmidrule(l){3-5}
    Alignment & Early stop & Clean & Randomized & Avg. \\
    \midrule
    None           & $\times$ & 54.38 & 8.75 & 31.56 \\
    Global         & $\times$ & 54.13 & 11.63 & 32.88 \\
    Patch-to-patch & $\times$ & \textbf{60.63} & 11.13 & 35.88 \\
    Patch-to-patch & $\checkmark$ & 59.13 & \textbf{13.25} & \textbf{36.19} \\
    \bottomrule
  \end{tabular*}
\end{table}

\subsection{Real-World Experiments}
\label{sec:bava_real_world}
\noindent\textbf{Experimental setting.} We conduct real-robot experiments on the bimanual AlohaMini2 platform. Each arm consists of a 6-DoF manipulator and a 1-DoF gripper, and the policy observes a head camera and two wrist cameras. We design three tasks: \emph{pick-and-place}, where the robot places a red block in a box; \emph{block stacking}, where it stacks a yellow block on a red one; and \emph{bimanual drawer placement}, where it opens a drawer with its left arm and places a block inside with its right. For each task, we train SmolVLA on 50 demonstrations without alignment, with Spatial Forcing, and with BAVA. On these demonstrations, DINOv2 has a higher BSM (0.292) than the pretrained SmolVLA (0.274), whereas VGGT has a lower one (0.163); BAVA therefore aligns to DINOv2, while Spatial Forcing aligns to VGGT. For evaluation, we run 10 trials per task and policy in a Clean setting that matches the data-collection environment and 10 in a Randomized setting with unseen tablecloth textures and distractors.

\begin{figure*}[t]
  \centering
  \includegraphics[width=\textwidth]{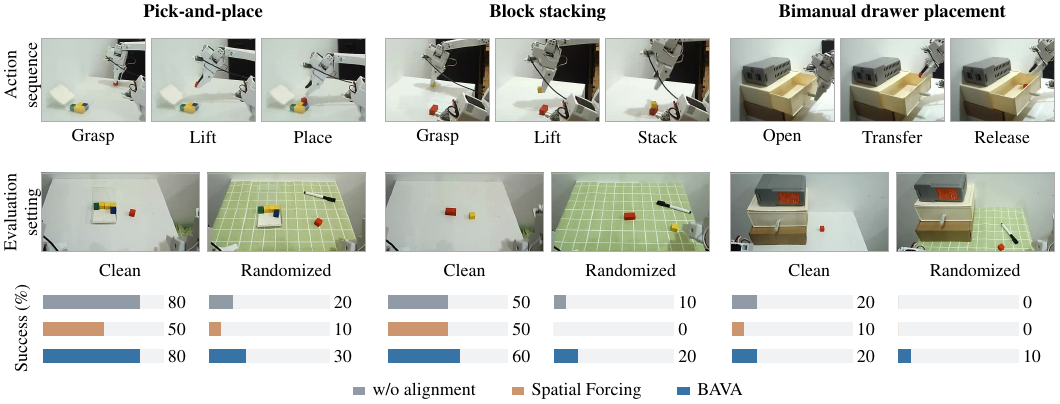}
  \caption{\textbf{Real-world manipulation on AlohaMini2.} \textit{Top:} Three keyframes from an expert demonstration of each task. \textit{Middle:} The Clean and Randomized evaluation settings. \textit{Bottom:} SmolVLA success rates (\%) over 10 trials per task, method, and setting.}
  \label{fig:real_robot}
\end{figure*}

\noindent\textbf{Evaluation results.} Across 60 trials per policy, BAVA succeeds in 22 (36.7\%), compared with 18 (30.0\%) for unaligned fine-tuning and 12 (20.0\%) for Spatial Forcing (\cref{fig:real_robot}). BAVA matches or exceeds unaligned fine-tuning in all six task--setting pairs, raising the Clean success rate from 50.0\% to 53.3\% and the Randomized one from 10.0\% to 20.0\%. Spatial Forcing, whose teacher has a lower BSM than the student, improves none of them.

\section{Related Work}
\label{sec:related}
\paragraph{Representation alignment in image generation.}
REPA aligns intermediate diffusion-transformer features with a pretrained vision encoder, substantially accelerating convergence and improving generation quality~\cite{yu2025repa}. Subsequent work investigates when this is effective. HASTE finds that alignment benefits can fade or even reverse late in training and proposes stopping alignment partway through training~\cite{wang2025haste}. iREPA compares 27 encoders and finds that spatial structure among patches, rather than global semantics, determines a teacher's value~\cite{singh2026irepa}. These studies select teachers using properties measured on the teacher alone before training. We extend this perspective to VLAs and identify a third property: behavioral structure relative to the student.

\paragraph{Representation alignment in VLAs.}
A series of methods uses frozen vision foundation model features as auxiliary training supervision and removes the teacher at inference time. These methods fall into three broad categories according to the prior they introduce. The first injects 3D spatial information. Spatial Forcing aligns visual tokens from one VLA layer with VGGT features~\cite{li2026spatialforcing}. ROCKET aligns multiple VLA layers with multiple layers of a 3D foundation model and uses shared projection heads to reduce gradient conflicts among layers of supervision~\cite{sun2026rocket}. RIGID aligns attention maps~\cite{zhu2026rigid}. The second category injects affordance priors. AffordVLA uses a frozen zero-shot affordance teacher to extract task-relevant visual representations from images and instructions, supervising intermediate VLA visual features~\cite{kong2026affordvla}. The third seeks to preserve pretrained representations. BlindVLA finds that action-only fine-tuning degrades VLA visual representations and anchors intermediate visual features to a frozen general-purpose vision encoder~\cite{kachaev2026blindvla}. MAPS instead adjusts module-specific constraints on the distance between parameters and their pretrained values~\cite{huang2025maps}. Although these approaches use different teachers and alignment formulations, they all assume that external teachers or pretrained representations provide something the student lacks. Their evaluations primarily rely on success rates on clean or nearly saturated benchmarks. We instead ask when this assumption holds, and evaluate under perturbations.

\paragraph{Visual representations for control.}
State abstraction groups states that are behaviorally equivalent; bisimulation metrics relax this equivalence into a distance between states~\cite{zhang2021learning}, and the policy similarity metric replaces reward differences with differences between optimal policies~\cite{agarwal2021pse}. A parallel line of work pretrains visual encoders for robot control and evaluates them as frozen policy backbones~\cite{nair2022r3m,majumdar2023vc1}, asking which encoder properties predict generalization under visual shifts~\cite{burns2024what}. BSM applies the behavioral-equivalence principle to a different question: rather than serving as a training objective or a backbone benchmark, it diagnoses whether a frozen encoder is a suitable alignment teacher for a given student.

\section{Conclusion}
\label{sec:conclusion}
In this paper, we revisited representation alignment for VLAs and asked what makes a teacher worth aligning to. Through a controlled study across teachers, students, and alignment methods, we found that off-the-shelf vision encoders bring little benefit to a strong pretrained VLA and can substantially reduce its robustness, and that the criteria used to select teachers in image generation do not explain these outcomes. To explain them, we introduced the behavioral structure metric (BSM), which measures whether states that are close in a teacher's feature space are followed by similar expert actions, and showed that alignment helps when the teacher's behavioral structure is richer than the student's. Building on this finding, we proposed BAVA, a simple recipe that selects teachers by BSM and stops alignment once the student's BSM plateaus. Simulation and real-world experiments show that BAVA improves both performance and robustness while aligning for only a fraction of training. We hope this study encourages choosing teachers for VLA alignment by the behavioral structure they encode.

\noindent\textbf{Limitations and future work.} Due to limited resources, our large-scale validation is conducted mainly in simulation, and our real-world validation remains limited. In future work, we plan to extend it to multiple robot embodiments and a wider range of tasks. Finally, this work uses BSM only as a diagnostic for selecting teachers; using it directly as a training objective for vision encoders would be a direction worth exploring.

\section*{LLM Usage Statement}
We used OpenAI Codex and Anthropic Claude Code to help modify code, analyze experimental results, write figure-plotting scripts, and polish the writing. All resulting code, analyses, figures, and text were verified by the authors.

\clearpage
\phantomsection
\addcontentsline{toc}{section}{References}
\bibliography{references}
\ifincludeappendix
  \clearpage
  \section*{Supplementary Material}
  \appendix
\section{Teachers}
\label{sec:teachers}
\noindent\textbf{Off-the-shelf vision encoders.} For each encoder, we extract features from the patch tokens of its final layer. We group the 11 encoders by how they are trained.

\emph{Image self-supervision.} DINOv2~\cite{oquab2024dinov2} is trained by self-distillation with image-level and masked patch-level objectives on 142M curated images. MAE~\cite{he2022mae} is trained to reconstruct masked patches of ImageNet-1K images.

\emph{Image--text alignment.} SigLIP~\cite{zhai2023siglip} is trained on image--text pairs with a pairwise sigmoid loss.

\emph{Segmentation and object understanding.} SAM3~\cite{carion2026sam3} is trained for promptable concept segmentation, in which it detects, segments, and tracks all objects matching a noun phrase or an image exemplar. We use its image detector with a text prompt for each task, and take the final layer of the encoder that fuses the image tokens with the prompt.

\emph{3D geometry.} VGGT~\cite{wang2025vggt} is a feed-forward transformer trained to predict camera parameters, depth maps, point maps, and point tracks from one or more views; we take the final layer of its multi-view aggregator. DA3~\cite{lin2026da3} is trained to predict depth and ray maps from an arbitrary number of views, using a plain transformer backbone.

\emph{Multi-teacher distillation.} RADIO~\cite{ranzinger2024radio} distills several vision foundation models into a single backbone. We use C-RADIOv3-L, a later release of this model family.

\emph{Video self-supervision.} V-JEPA~2.1~\cite{murlabadia2026vjepa21} is trained on images and videos with a masked latent-prediction objective in which both visible and masked tokens contribute to the loss. DINO-Video~\cite{wu2026lingbotvla2} extends a pretrained image backbone with causal temporal attention and is trained with video-adapted DINO and iBOT objectives on 5M video clips spanning internet, egocentric, and robot data.

\emph{Embodied visual control.} CAIP~\cite{sharma2026caip} is trained with a contrastive objective between images and actions, using 3D hand keypoints from egocentric human video as a proxy for end-effector actions. VC-1~\cite{majumdar2023vc1} is trained by masked autoencoding on over 4,000 hours of egocentric video together with ImageNet.

\noindent\textbf{Reference teachers.} Random ViT has the architecture of DINOv2 with randomly initialized, frozen weights, and uses the same feature readout. The fine-tuned $\pi_0$ teacher is the pretrained $\pi_0$ fine-tuned with the action loss alone for 50k iterations on the student's 800 Clean demonstrations. The oracle is a $\pi_{0.5}$ policy fine-tuned with the action loss alone for 20k iterations on expert demonstrations that cover the Randomized setting, 550 per task and 8,800 in total. For both policy teachers, we use the visual tokens at the 12th block of the VLM.

\section{Tasks}
\label{sec:tasks}
We use 16 manipulation tasks from RoboTwin~2.0~\cite{chen2025robotwin2}, grouped into four categories of four tasks each. \Cref{fig:tasks_overview} shows three keyframes from an expert demonstration of each task.

\begin{figure*}[p]
  \begin{minipage}[c][\textheight][c]{\textwidth}
  \centering
  \includegraphics[width=\textwidth]{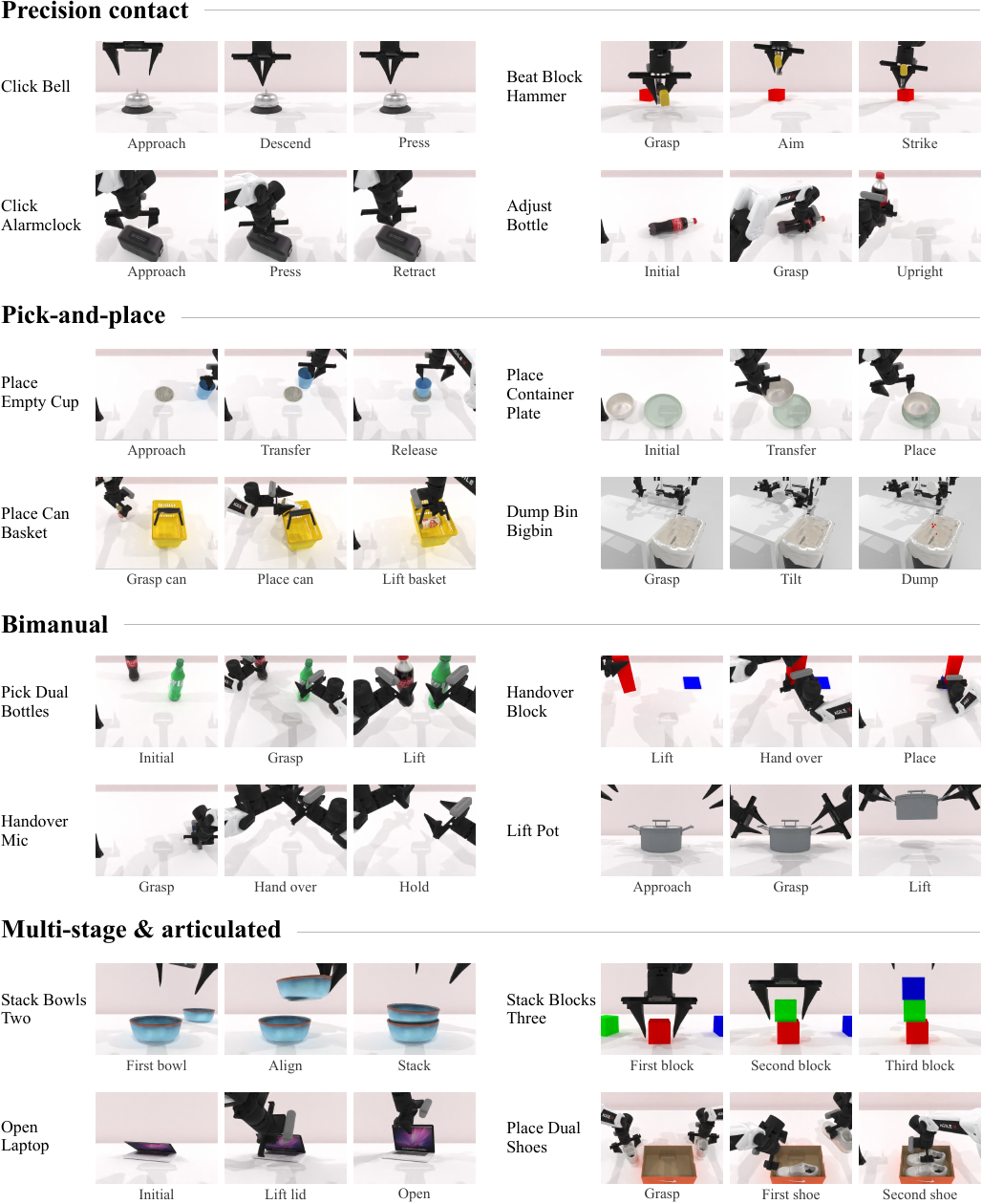}
  \caption{\textbf{The 16 RoboTwin tasks.} Three keyframes from an expert demonstration of each task, grouped by category.}
  \label{fig:tasks_overview}
  \end{minipage}
\end{figure*}

\noindent\textbf{Precision contact.} The robot reaches a small target and controls where it makes contact or how it grasps.
\begin{itemize}[nosep,leftmargin=1.2em]
  \item \emph{Click Bell}: Press the button on top of a desk bell.
  \item \emph{Beat Block Hammer}: Pick up a hammer and strike a block.
  \item \emph{Click Alarmclock}: Press the button on top of an alarm clock.
  \item \emph{Adjust Bottle}: Pick up a bottle lying on its side and turn it upright.
\end{itemize}

\noindent\textbf{Pick-and-place.} The robot picks up an object and moves it onto a support or into a container.
\begin{itemize}[nosep,leftmargin=1.2em]
  \item \emph{Place Empty Cup}: Place a cup on a coaster.
  \item \emph{Place Container Plate}: Place a container on a plate.
  \item \emph{Place Can Basket}: Place a can in a basket and lift the basket.
  \item \emph{Dump Bin Bigbin}: Empty a small bin into a large bin beside the table.
\end{itemize}

\noindent\textbf{Bimanual.} The two arms lift objects together or hand an object from one to the other.
\begin{itemize}[nosep,leftmargin=1.2em]
  \item \emph{Pick Dual Bottles}: Grasp a bottle with each gripper and lift both.
  \item \emph{Handover Block}: Pass a block from one arm to the other and place it on a target pad.
  \item \emph{Handover Mic}: Pick up a microphone and pass it to the other arm.
  \item \emph{Lift Pot}: Grasp the two handles of a pot and lift it.
\end{itemize}

\noindent\textbf{Multi-stage and articulated.} The robot completes several sub-goals in sequence or manipulates a hinged object.
\begin{itemize}[nosep,leftmargin=1.2em]
  \item \emph{Stack Bowls Two}: Stack one bowl on another.
  \item \emph{Stack Blocks Three}: Stack three blocks into a tower.
  \item \emph{Open Laptop}: Open a laptop by lifting its screen.
  \item \emph{Place Dual Shoes}: Place two shoes side by side in a shoebox.
\end{itemize}

\section{Details of BSM}
\label{sec:bsm_details}
\subsection{Definition}
We give the complete definition of BSM.

\noindent\textbf{State feature.} For each camera $c$, we average the patch features into $\bar{\mathbf{f}}_i^{\,c}$ and normalize it to unit length. The state feature concatenates the three cameras:
\begin{equation*}
\mathbf{z}_i=\left[\frac{\bar{\mathbf{f}}_i^{\,1}}{\lVert\bar{\mathbf{f}}_i^{\,1}\rVert};\ \frac{\bar{\mathbf{f}}_i^{\,2}}{\lVert\bar{\mathbf{f}}_i^{\,2}\rVert};\ \frac{\bar{\mathbf{f}}_i^{\,3}}{\lVert\bar{\mathbf{f}}_i^{\,3}\rVert}\right].
\end{equation*}
The neighbors $\mathcal{N}_k^\tau(i)$ are the $k$ other states of the same task with the highest cosine similarity to $\mathbf{z}_i$.

\noindent\textbf{Action chunk.} We follow the ALOHA configuration of openpi, the official implementation of $\pi_0$ and $\pi_{0.5}$~\cite{black2025pi0,pi2025pi05}. For a state $i$ at time $t$, the action chunk consists of the next $h$ expert steps, $\mathbf{a}_{i,1},\dots,\mathbf{a}_{i,h}$. Each step is a 14-dimensional vector that contains, for each arm, the six joint positions relative to the current state, $\mathbf{q}_{t+n}-\mathbf{q}_t$, and the gripper opening. Each dimension $d$ is normalized with the 1st and 99th percentiles of the training actions, $Q_{0.01,d}$ and $Q_{0.99,d}$:
\begin{equation*}
\tilde a_{i,n,d}=2\,\frac{a_{i,n,d}-Q_{0.01,d}}{Q_{0.99,d}-Q_{0.01,d}+10^{-6}}-1.
\end{equation*}
For a state with fewer than $h$ remaining steps, the last step is repeated.

\noindent\textbf{Behavioral distance.} The behavioral distance between two states is the root-mean-square difference between their normalized action chunks:
\begin{equation*}
d_B(i,j)=\sqrt{\frac{1}{14h}\sum_{n=1}^{h}\sum_{d=1}^{14}\bigl(\tilde a_{i,n,d}-\tilde a_{j,n,d}\bigr)^2}.
\end{equation*}

\noindent\textbf{BSM.} With the neighbors and the behavioral distance defined above, BSM in each setting is
\begin{equation*}
\bsm=\frac{1}{|\mathcal{T}|}\sum_{\tau\in\mathcal{T}}\left[
1-\frac{1}{|\mathcal{S}_\tau|}\sum_{i\in\mathcal{S}_\tau}
\frac{\frac{1}{k}\sum_{j\in\mathcal{N}_k^\tau(i)}d_B(i,j)}
{\mathbb{E}_{j\in\mathcal{S}_\tau,\,j\ne i}\,d_B(i,j)}
\right],
\end{equation*}
where $\mathcal{T}$ is the set of tasks and $\mathcal{S}_\tau$ the states sampled from task $\tau$. We use $k=10$, $h=50$, and 100 states per task.

\subsection{Sensitivity to the Metric Definition}
\noindent\textbf{Neighbors and horizon.} We recompute BSM for all 14 teachers with $k\in\{5,10,20\}$ neighbors and action horizons of $h\in\{10,20,50\}$ steps. \Cref{tab:kh_sensitivity} reports Pearson's $r$ between BSM and success rate in the Randomized setting for each combination. The correlation stays between 0.82 and 0.89 and is significant in every case ($p<0.001$).

\begin{table}[t]
  \centering
  \caption{\textbf{Sensitivity of the correlation to $k$ and $h$.} Pearson's $r$ between BSM and success rate in the Randomized setting across the 14 teachers. The main text uses $k=10$ and $h=50$.}
  \label{tab:kh_sensitivity}
  \small
  \setlength{\tabcolsep}{2pt}
  \begin{tabular*}{.53\linewidth}{@{\extracolsep{\fill}}lccc@{}}
    \toprule
    & $h=10$ & $h=20$ & $h=50$ \\
    \midrule
    $k=5$ & 0.87 & 0.87 & 0.89 \\
    $k=10$ & 0.85 & 0.85 & 0.87 \\
    $k=20$ & 0.82 & 0.82 & 0.84 \\
    \bottomrule
  \end{tabular*}
\end{table}

\noindent\textbf{Behavioral distance.} We change the behavioral distance in two ways, keeping everything else fixed: we represent actions as end-effector poses instead of joint positions, or replace the root-mean-square difference with the mean absolute difference. In both cases, the correlation with Randomized success rate remains $r=0.87$, and the ranking of the 14 teachers by BSM is unchanged.

\noindent\textbf{Feature similarity.} For simplicity, BSM compares states by their pooled features. We also test a patch-level similarity: the cosine similarity between patch features at the same position, averaged over all patches and cameras. With this similarity, the correlation with Randomized success rate is $r=0.80$ and remains highly significant ($p<0.001$).

\section{Implementation Details}
\label{sec:implementation_details}
\subsection{Alignment Details}
We describe the four alignment methods compared in \cref{tab:methods}. All four align $\pi_{0.5}$ to VGGT with $\lambda=0.1$ throughout training, and teacher features are resized to the student's token grid by bilinear interpolation.

\emph{Patch-to-patch.} The student's visual tokens at the 12th block pass through an MLP and are matched to the final-layer VGGT features with a cosine loss averaged over patches. The MLP consists of layer normalization, a linear layer, GELU, and a second linear layer, with a width of 2,048.

\emph{Attention-map.} At the 12th block, two learned linear projections map the student's visual tokens to queries and keys. The resulting attention map is matched to the global attention of the 11th VGGT block, averaged over heads, with an entropic optimal transport loss computed for each pair of views.

\emph{Multi-layer.} Ten layer pairs are aligned with equal weights: student blocks 1 to 9 and 17 are paired in order with VGGT blocks 2, 4, \dots, 18, and 23. Each pair uses the patch-to-patch loss, and all pairs share one Matryoshka MLP.

\emph{Global.} For each camera, the student's visual tokens at the 12th block and the final-layer VGGT features are averaged over patches. The student's averages pass through the same MLP as in patch-to-patch alignment, and the cosine loss is averaged over the three cameras.

\begin{table*}[t]
  \centering
  \caption{\textbf{Teacher metrics and alignment outcomes for all 14 teachers}, sorted by BSM in the Randomized setting. LP is ImageNet-1K linear probing accuracy (\%). Success rates (\%) are those of the $\pi_{0.5}$ student aligned patch-to-patch to each teacher, at 30k iterations over 800 episodes per setting; parentheses give the difference from unaligned fine-tuning. The first row gives the success rates of the student fine-tuned without alignment and the BSM of the pretrained $\pi_{0.5}$ before fine-tuning.}
  \label{tab:teacher_metrics}
  \small
  \setlength{\tabcolsep}{3pt}
  \begin{tabular*}{\textwidth}{@{\extracolsep{\fill}}llrrrrrrr@{}}
    \toprule
    & & & \multicolumn{2}{c}{LDS} & \multicolumn{2}{c}{BSM} & \multicolumn{2}{c}{Success rate (\%)} \\
    \cmidrule(lr){4-5}\cmidrule(lr){6-7}\cmidrule(l){8-9}
    Teacher & Family & LP & Clean & Rand. & Clean & Rand. & Clean & Randomized \\
    \midrule
    w/o alignment & -- & -- & -- & -- & 0.591 & 0.497 & 84.00 & 58.13 \\
    \midrule
    Oracle & Reference & 57.5 & 0.03 & 0.03 & 0.670 & 0.669 & 86.88 (+2.88) & 83.13 (+25.00) \\
    Fine-tuned $\pi_0$ & Reference & 37.6 & 0.03 & 0.02 & 0.669 & 0.636 & 82.88 ($-$1.13) & 49.00 ($-$9.13) \\
    DINOv2 & Image SSL & 82.6 & 0.27 & 0.31 & 0.562 & 0.457 & 81.88 ($-$2.13) & 36.38 ($-$21.75) \\
    RADIO & Distillation & 83.2 & 0.21 & 0.22 & 0.554 & 0.444 & 83.25 ($-$0.75) & 34.38 ($-$23.75) \\
    SAM3 & Segmentation & 53.5 & 0.43 & 0.44 & 0.523 & 0.423 & 81.00 ($-$3.00) & 41.50 ($-$16.63) \\
    SigLIP & Image--text & 83.7 & 0.14 & 0.15 & 0.534 & 0.407 & 78.13 ($-$5.88) & 36.13 ($-$22.00) \\
    V-JEPA 2.1 & Video SSL & 76.0 & 0.27 & 0.25 & 0.569 & 0.401 & 84.25 (+0.25) & 33.25 ($-$24.88) \\
    VGGT & 3D geometry & 58.5 & 0.09 & 0.11 & 0.449 & 0.378 & 83.88 ($-$0.13) & 47.75 ($-$10.38) \\
    DINO-Video & Video SSL & 74.7 & 0.16 & 0.17 & 0.557 & 0.357 & 82.63 ($-$1.38) & 38.63 ($-$19.50) \\
    DA3 & 3D geometry & 38.2 & 0.03 & 0.04 & 0.363 & 0.317 & 80.38 ($-$3.63) & 42.75 ($-$15.38) \\
    CAIP & Embodied & 39.0 & 0.29 & 0.30 & 0.556 & 0.291 & 81.00 ($-$3.00) & 30.13 ($-$28.00) \\
    VC-1 & Embodied & 63.8 & 0.34 & 0.48 & 0.524 & 0.177 & 80.63 ($-$3.38) & 19.63 ($-$38.50) \\
    MAE & Image SSL & 66.7 & 0.25 & 0.42 & 0.529 & 0.133 & 81.00 ($-$3.00) & 19.13 ($-$39.00) \\
    Random ViT & Reference & 6.8 & 0.11 & 0.12 & 0.495 & 0.080 & 81.25 ($-$2.75) & 7.13 ($-$51.00) \\
    \bottomrule
  \end{tabular*}
\end{table*}
\subsection{Training}
All four students are trained jointly on the 16 tasks, with 50 Clean demonstrations per task.

\emph{$\pi_{0.5}$.} We train with openpi from the pretrained base checkpoint and update all parameters. We use AdamW with a batch size of 64. The learning rate is warmed up to $5\times10^{-5}$ over 1,000 iterations and then follows a cosine decay to $5\times10^{-6}$ at 30k iterations.

\emph{$\pi_0$.} We use the same setting as for $\pi_{0.5}$, starting from the pretrained $\pi_0$ base checkpoint.

\emph{OpenVLA-OFT.} We initialize from the pretrained OpenVLA weights and train LoRA adapters of rank 32 together with a continuous action head that predicts chunks of 25 actions. We use AdamW with a batch size of 64 and a constant learning rate of $5\times10^{-4}$ for 50k iterations.

\emph{SmolVLA.} We initialize from the pretrained SmolVLA checkpoint and update both the VLM and the action expert. We use AdamW with a batch size of 64 and a learning rate of $1\times10^{-4}$ for 30k iterations.

\subsection{Real-World Experiments}
\label{sec:real_world_details}
\emph{Training.} For each task, we train one SmolVLA policy per method on 50 demonstrations for 50k iterations, aligning at the same layer as in simulation with $\lambda=0.1$. Spatial Forcing aligns to VGGT throughout training, and BAVA aligns to DINOv2 and turns off alignment at 10k iterations.

\begin{table}[t]
  \centering
  \caption{\textbf{One teacher, four students.} All students are aligned patch-to-patch to DINOv2 (BSM 0.509). BSM is that of the pretrained student, averaged over the Clean and Randomized settings; $\Delta$BSM is the teacher's BSM minus the student's. $\Delta$SR is the change in average success rate relative to unaligned fine-tuning.}
  \label{tab:same_teacher_students}
  \small
  \setlength{\tabcolsep}{2pt}
  \begin{tabular*}{.63\linewidth}{@{\extracolsep{\fill}}lrrr@{}}
    \toprule
    Student & BSM & $\Delta$BSM & $\Delta$SR (pp) \\
    \midrule
    OpenVLA-OFT & 0.314 & $+0.195$ & $+8.25$ \\
    SmolVLA & 0.459 & $+0.050$ & $+4.31$ \\
    $\pi_0$ & 0.522 & $-0.013$ & $-4.69$ \\
    $\pi_{0.5}$ & 0.544 & $-0.035$ & $-11.94$ \\
    \bottomrule
  \end{tabular*}
\end{table}

\emph{BSM.} We compute BSM on 20 separate demonstrations per task, sampling 10 states at equal intervals from each. We use $k=10$ neighbors and a horizon of $h=25$ steps, which corresponds to one second. Each action dimension is normalized by its mean and standard deviation, and neighbors are retrieved from the other demonstrations of the same task.

\emph{Evaluation.} A trial counts as a success when the task goal is reached:
\begin{itemize}[nosep,leftmargin=1.2em]
  \item \emph{Pick-and-place}: The red block is inside the box.
  \item \emph{Block stacking}: The yellow block rests on top of the red block.
  \item \emph{Bimanual drawer placement}: The drawer is opened with the left arm and the block is placed inside with the right arm.
\end{itemize}
In the Randomized setting, the table is covered with an unseen tablecloth, and the distractors are rearranged every two trials.

\section{Complete Results}
\label{sec:full_alignment_results}
\subsection{Teacher Metrics and Alignment Outcomes}
\Cref{tab:teacher_metrics} reports the three teacher metrics and the alignment outcome for each of the 14 teachers studied in \cref{sec:behavioral_structure}, together with the $\pi_{0.5}$ student. Every aligned policy uses patch-to-patch alignment and is trained for 30k iterations.

\begin{figure*}[t]
  \centering
  \includegraphics[width=\textwidth]{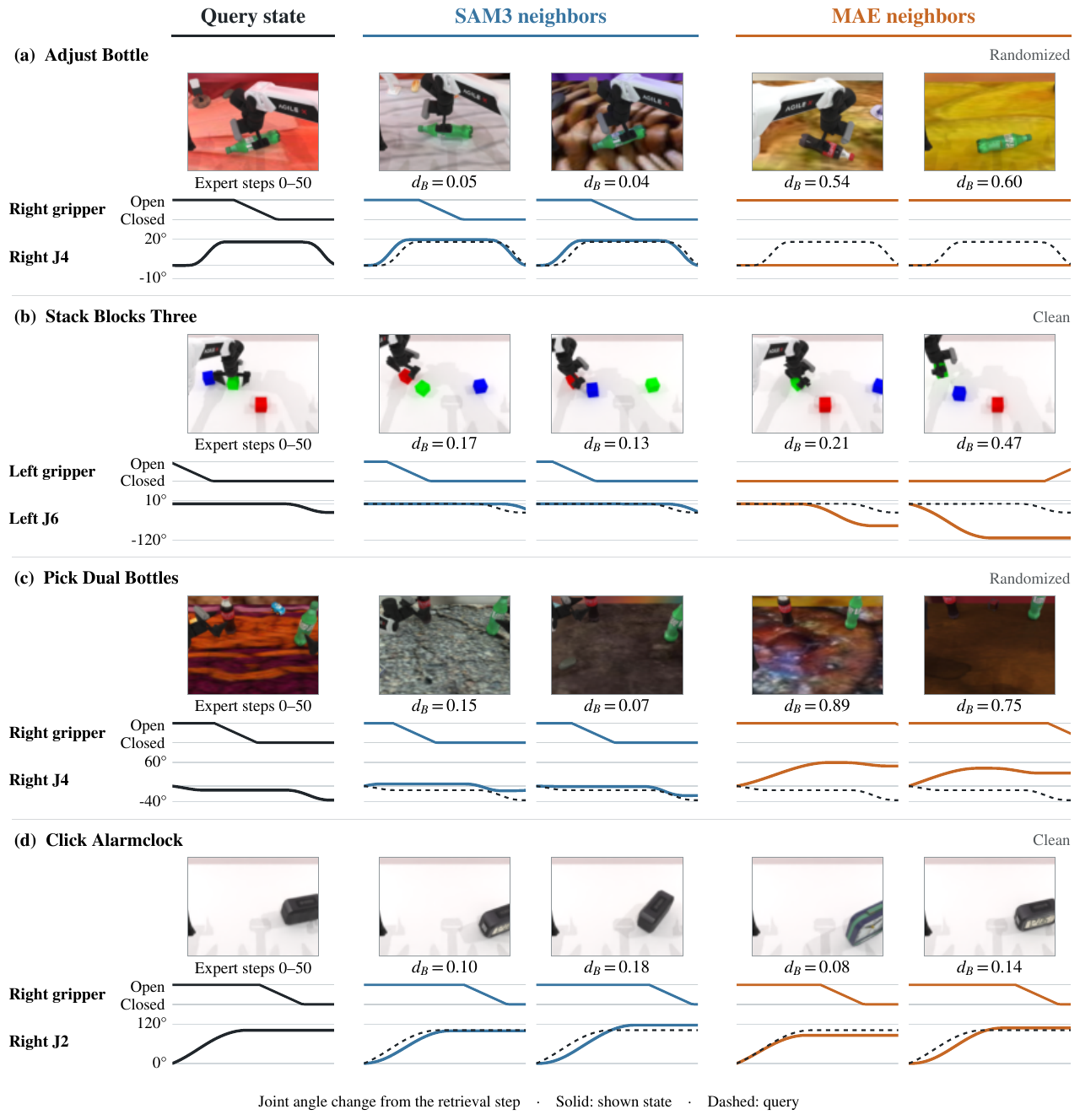}
  \caption{\textbf{Retrieval examples for SAM3 and MAE.} Each row shows a query state and its two nearest neighbors in the feature space of SAM3 (blue) and MAE (orange), with the head-camera view at the retrieval step. The traces below each image plot the gripper opening and the change of one joint angle over the next 50 expert steps. The joint is the one that moves farthest in the query, and dashed lines repeat the query's trajectory. $d_B$ is the behavioral distance to the query. SAM3's neighbors follow the query's actions in all four rows, whereas MAE's do so only in (d).}
  \label{fig:appendix_retrieval_examples}
\end{figure*}
\subsection{Alignment across Students}
\label{sec:other_students}
\Cref{tab:same_teacher_students} collects the outcome of aligning four students to the same teacher, DINOv2, with patch-to-patch alignment throughout training. The two students whose BSM is lower than that of DINOv2 improve, the two whose BSM is higher degrade, and the change in success rate decreases monotonically with the student's BSM.

\subsection{Additional Retrieval Examples}
\Cref{fig:appendix_retrieval_examples} shows retrieval examples for SAM3 and MAE, whose BSM in the Randomized setting is 0.423 and 0.133. For each query, we show the two nearest neighbors in each feature space, together with the expert's gripper and joint motion over the next 50 steps. In (a)--(c), the states retrieved by SAM3 are followed by nearly the same motion as the query, while those retrieved by MAE are not. In (d), both encoders retrieve states with similar actions.

\fi
\end{document}